\documentclass[11pt,a4paper]{llncs}
\usepackage{amsmath}
\usepackage{amssymb}
\usepackage{graphicx}
\usepackage{marvosym}
\usepackage{url}
\usepackage{fancyhdr}
\usepackage{hyperref}

\usepackage{geometry}
\fancypagestyle{firstpage}{\fancyhf{}
}

\usepackage{pgfplots}
\pgfplotsset{compat=1.17}

\begin{document}


\title{\LARGE{Mixed-Distil-BERT: Code-mixed Language Modeling for Bangla, English, and Hindi}}


%
%
\author{\large{Md Nishat Raihan  \and Dhiman Goswami \and Antara Mahmud}}
\institute{\large{George Mason university\\ Fairfax, Virginia, USA \\ mraihan2@gmu.edu, dgoswam@gmu.edu, amahmud4@gmu.edu}}

%


%
%


\maketitle

\pagestyle{plain}

\thispagestyle{firstpage}

\begin{abstract}
One of the most popular downstream tasks in the field of Natural Language Processing is text classification. Text classification tasks have become more daunting when the texts are code-mixed. Though they are not exposed to such text during pre-training, different BERT models have demonstrated success in tackling Code-Mixed NLP challenges. Again, in order to enhance their performance, Code-Mixed NLP models have depended on combining synthetic data with real-world data. It is crucial to understand how the BERT models' performance is impacted when they are pretrained using corresponding code-mixed languages. In this paper, we introduce Tri-Distil-BERT, a multilingual model pre-trained on Bangla, English, and Hindi, and Mixed-Distil-BERT, a model fine-tuned on code-mixed data. Both models are evaluated across multiple NLP tasks and demonstrate competitive performance against larger models like mBERT and XLM-R. Our two-tiered pre-training approach offers efficient alternatives for multilingual and code-mixed language understanding, contributing to advancements in the field. \textit{Both models are available at huggingface. \footnote{https://huggingface.co/md-nishat-008/Tri-Distil-BERT} \footnote{https://huggingface.co/md-nishat-008/Mixed-Distil-BERT}}
\end{abstract}


\section{Introduction}
Code-mixing and Code-switching are very common linguistic cases observed both in speech and text form. Code-mixing means the practice of using words from multiple languages within a single utterance, sentence, or discourse, and Code-switching refers to the deliberate alteration between multiple languages within the same context \cite{thara2018code}. The first case is spontaneous and the second case is purposeful. However, both are widely observed in bilingual and multilingual communities. 

Though code-mixed texts are mostly common in verbal speeches, in this era of social media, people highly tend to use a mixture of multiple languages in their posts and comments. Again, such contents can be a great source of information for various research purposes. Multilingual code-mixed sentence labeling is important for applications such as sentiment analysis, named entity recognition, and machine translation in multilingual regions and online communication platforms. In this paper, we are working on sentiment analysis, offensive language detection, and multi-label emotion classification. For each of these tasks, we have generated three datasets, each of which is three language (English, Bengali, and Hindi) code-mixed datasets augmented from social media posts. These datasets might be proven to be very helpful for further research in this direction. 

The three downstream tasks, we are trying to address - sentiment analysis, offensive language detection, and multi-label emotion classification are well-known problems in the field of natural language processing. But when code-mixing is taken into consideration, these tasks become computationally challenging. Previous researches have been performed on two languages of code-mixing for the mentioned tasks. However, as per our knowledge, till now, no one explored these tasks' efficacy for code-mixed data incorporating three languages.  A strong reason to come up with the motivation of generating a dataset from code-mixed social media posts is that people use code-mixed languages only for informal purposes and social media is the best place to encounter code-mixed texts. No formal datasets are available to work on. 

At first, we have pre-trained a trilingual DistilBERT in Bangla, and Hindi language. Then, we pre-train this trilingual DistilBERT with code-mixed data in English-Bengali-Hindi. Then we finetune our two pre-trained DistilBERT models along with other BERT models for sentiment analysis, offensive language detection, and multi-label emotion detection. Among the other BERT models, we include BERT, Bangla-BERT, Indic-BERT, mBERT, robERTa, xlm-roBERTa, ro-BERTa, Banglish-BERT, Hing-BERT etc. 

Our results indicate that our two pre-trained DistilBERT models perform quite satisfactorily and similarly to other conventional BERT models for all three tasks. Our observation indicates that the performance of Tri-Distil-BERT is comparable to mBERT whereas the performance of Mixed-Distil-BERT can be compared to other two-level code-mixed BERT models like BanglishBERT and HingBERT. However, in some cases, Mixed-Distil-BERT outperforms the abovementioned two BERT models.

\section{Background and Related Works}

\subsection{Background on Important or Non-Standard Concepts}
\textbf{Code Mixing:} Code-Mixing is the process of incorporating linguistic building blocks such as words, phrases, and morphemes, from one language  into an expression from another. It is a common practice of using multiple languages in a single utterance on social media sites like Facebook, WhatsApp, etc \cite{thara2018code}.\\
Example: "Oidin amar old school e gesilam. Teacher der dekhe ja good feeling hoise!"\\\\
\textbf{Difference between bilingual and tri-lingual code-mixing:} Bilingual code-mixing involves combining language blocks of two languages in a single sentence. Whereas, tri-lingual code-mixing refers to the process of mixing linguistic blocks of three languages into one sentence.\\
Example of Bilingual code-mixing: "Park e giye first din ei ato baaje experience hoise."\\
Example of Tri-lingual code-mixing: "Jibone ato saddness, accha nehi feel hoy."\\\\
\textbf{Pre-training:} Pretraining is referred to as a training method where a model's weights are trained on a big dataset to be used as a starting point for training on smaller, related datasets.\\\\
\textbf{Transliteration:} A mapping from one writing system into another, usually from grapheme to grapheme, is known as transliteration \cite{weinberg1974transliteration}. Texts are called transliterated when they are changed from one script to another based on phonetic similarities.\\
Example: "Ami tomak valobashi" is an English transliteration of a Bengali sentence whose English Translation is: "I love you."\\\\
\textbf{Perplexity:} How confidently a language model predicts a text sample is measured statistically by perplexity. In other words, it measures the degree to which a model is "surprised" by fresh data. The model can anticipate the text more accurately as perplexity decreases\cite{techslang}.

\subsection{Related work}
Code-mixing, or the use of multiple languages in one speech, is a common phenomenon in multilingual cultures. Natural language processing (NLP) models that can successfully handle code-mixed text are increasingly needed due to the prominence of code-mixed data in social media, online communication, and other digital platforms. \cite{chanda2016unraveling} proposed a system that performs language identification and morphological disambiguation in code-switching text. The system performs both tasks with great accuracy on a code-switching corpus of Hindi-English and Bengali-English text. The suggested approach contributes to the creation of code-switching language processing tools for a variety of uses, including machine translation, social media analysis, and language instruction.

Again, in \cite{barman2014code}, the authors have shared their ongoing research on the issue of code-mixed social media data and artificial language identification. They discuss a brand-new dataset made up of Facebook comments and posts that exhibit code-mixing between Bengali, English, and Hindi. Additionally, they provide some preliminary research on word-level language recognition using several approaches, such as an unsupervised dictionary-based approach, supervised word-level classification \cite{kushol2018android} with and without contextual cues, and conditional random fields for sequence labeling. Similar works include \cite{goswami2024masontigers}, \cite{raihan2023nlpbdpatriots} and \cite{goswami2023nlpbdpatriots}. Some works also focus on words extracted from image data, which is very likely to be code mixed as well \cite{ahmed2019complete}. Such works also need to focus on traditional edge detection methods \cite{raihan2022experimental}.

On the other hand, code mixed data is difficult to find in bulk quantity as it is informal in nature and mostly available in social media posts only. The only relevant works include SentMix-3L \cite{raihan2023sentmix}, OffMix-3L \cite{goswami2023offmix} and TB-OLID \cite{raihan2023offensive}. This data shortage has already been addressed by creating synthetic code mixed data to supplement the real code mixed data. Parallel sentences can be used to create code-mixed sentences, as shown by \cite{pratapa2018word} and \cite{bhat2016grammatical}, who also illustrate how utilizing these artificial sentences can increase language model confusion. By employing phrase tables to align and combine components of a parallel sentence, \cite{yang2020alternating} creates CM sentences. \cite{samanta2019improved} also put out a similar approach that creates artificial phrases using parse trees. When optimizing a transformer for a downstream task, \cite{krishnan2021multilingual} provides a novel way to enhance the monolingual source data via multilingual code-switching via random translations.

While previous works have explored the performance of multilingual models on code-mixed text as we have seen above, the base paper, \cite{santy2021bertologicomix} goes further by analyzing how code-mixing impacts multilingual model performance specifically for English-Hindi and English-Spanish. They run a number of tests to assess how various multilingual BERT models perform on the code-mixed text and look into the impact of various variables. Also, they have run the experiments on a number of NLP tasks rather than focusing on only one type of task, as we have seen previously.

Inspired by the work done in \cite{santy2021bertologicomix}, we plan to analyze how code-mixing impacts multilingual model performance for a combination of three languages (Hindi, Bangla, English). Also, instead of trying out similar NLP tasks as the mentioned paper, we plan to attempt three distinct NLP tasks.

\subsection{Proposed Approach}

\begin{figure*}[!h]
    \centering
    \includegraphics[width=\columnwidth]{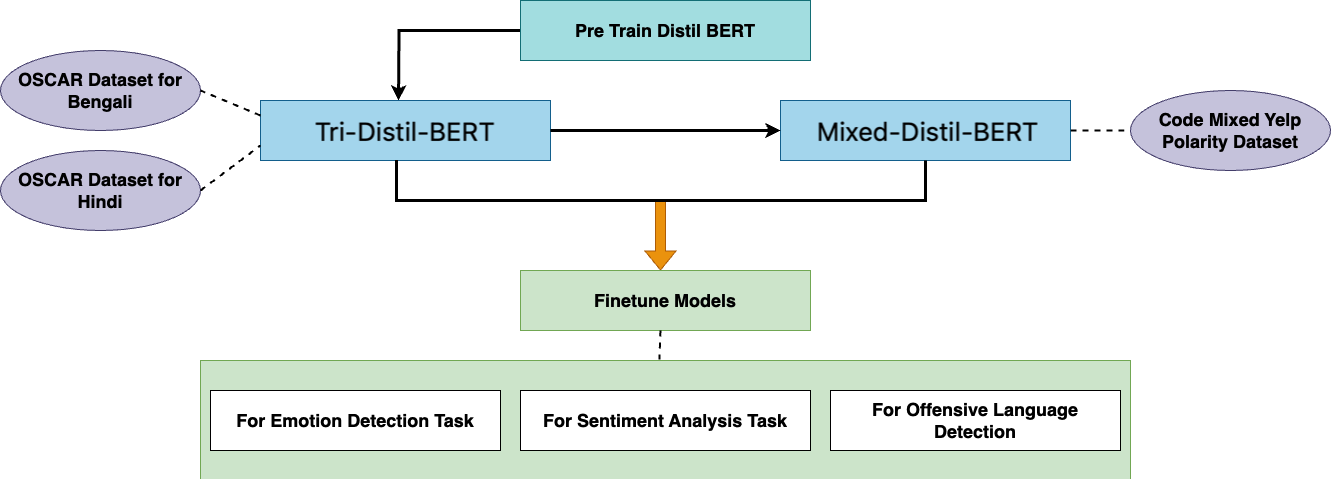}
    \caption{Workflow of the pre-trained models}
    \label{fig:flowchart}
    \end{figure*}
We have attempted to evaluate the performance of BERT models in two manners. Firstly we wanted to see if the claim of the base paper \cite{santy2021bertologicomix} holds for a tri-lingual code-mixed setting. Secondly, we wanted to explore how would the existing BERT models perform if they are pre-trained with tri-lingual code-mixed data instead of pure mono-lingual data.\\
We can divide our overall workflow into two phases. Figure \ref{fig:flowchart} depicts the flow of work. In the first phase, we pre-train one BERT model with tri-lingual code-mixed data. Originally our plan was to pre-train roBERTa model. However, due to computational constraints, we decided to go with DistilBERT, which is a compact, yet effective version of the BERT model. \\
We have done the pre-training in two ways. Firstly, pre-trained DistilBERT with the Bengali and Hindi forms of the OSCAR dataset. The details of the dataset is given in the later part of the report. We have named this model as Tri-Distil-BERT. Afterward, we have pre-trained this Tri-Distil-BERT model with a tri-linugal code-mixed dataset. We have named this version of the model as Mixed-Distil-BERT. In this case, we used the Yelp polarity dataset converted into a tri-lingual (English-Bengali-Hindi) code-mixed version.\\
In the second phase of our work, we have fine-tuned our two pre-trained models along with other BERT models to evaluate the performance of our newly pre-trained models. We have finetuned the models for three distinct NLP tasks: multi-label emotion detection, multi-label sentiment analysis, and offensive language detection. The dataset contains code-mixed synthetic data generated by us. \\
After finetuning, we compare the results and run our analysis to obtain insight into the comparative performance of the models. Also, we try to investigate the impact of the nature of the datasets on the models' performance to thoroughly understand what sort of effect code-mixed pre-training has on the performance of the models.\\

\section{Experiments}

\subsection{Datasets}

We have used the dataset Emotion detection from the text \href{https://www.kaggle.com/datasets/gangulyamrita/social-media-emotion-dataset}{SMED}\footnote{https://www.kaggle.com/datasets/gangulyamrita/social-media-emotion-dataset}  for multi-label emotion detection, Offensive Language Identification (OLID) \cite{zampieri2019semeval} and A Large-Scale Semi-Supervised Dataset for Offensive Language Identification (SOLID) \cite{rosenthal2020solid} for offensive language classification, and Amazon Review Data \cite{ni2019justifying} for sentiment analysis as seed for our synthetic datasets. Then we performed code-mixing for English, Bengali, and Hindi by random code-mixing algorithm \cite{krishnan2021multilingual} and generated a 3-language code-mixing data of 100k. For offensive language classification, we have used 14100 data from the OLID dataset and 85900 data from the SOLID dataset. For sentiment analysis, we have used 100k data from the Amazon Review Dataset. We have applied the same code-mixing approach that we performed for Emotion detection. For all the datasets train, val and test split will be 60\%, 20\%, and 20\%.

\subsection{Model Pre-train}
In the domain of code-mixing, there is no dedicated and standardized LLM. There are models that support multilinguality like mBERT, and XLM-R but these models also have limitations for code-mixing and transliterations. Moreover, some bi-lingual models like banglishBERT or HingBERT also tried to address the same issue. But in practice, people in the south Asian region normally use code-mixing of more than two languages. No existing languages model performs well on more than 2 languages and models like banglishBERT and HingBERT fail here. Moreover, the use of Bengali, Hindi, and English is also very common in the Asian subcontinent region. That is the reason we want to pre-train a Language model with Bengali, Hindi, and English language. We have chosen DistilBERT which is smaller, faster, and similar in performance in comparison to the BERT model. The DistilBERT model is already pretrained in the English language. We have used the Bengali and Hindi corpus of the OSCAR dataset \cite{suarez2020monolingual} to pretrain the DistilBERT model. From now on it will be referred as Tri-Distil-BERT. Now to introduce code-mixing to this pre-trained model, We have used yelp polarity dataset \cite{zhangCharacterlevelConvolutionalNetworks2015} and performed code-mixing for English, Bengali, and Hindi by random code-mixing algorithm \cite{krishnan2021multilingual}. Then we further pre-trained Tri-Distil-BERT with the code-mixed version of the Yelp polarity dataset. This pre-trained model will be referred to further as Mixed-Distil-BERT.\\

\begin{table*}
  \centering
  \footnotesize
  \setlength{\tabcolsep}{3.75pt}
  \renewcommand{\arraystretch}{1.5}
  \caption{Training and Validation Loss of our Pre-trained Models}
  \begin{tabular}{ | c | c | c | c |c | c | }
    \hline
    \multicolumn{3}{|c|}{\textbf{Tri-Distil-BERT}} & \multicolumn{3}{|c|}{\textbf{Mixed-Distil-BERT}} \\
    \cline{1-6}
    \textbf{Epoch} & \textbf{Training Loss} & \textbf{Validation Loss} & \textbf{Epoch} & \textbf{Training Loss} & \textbf{Validation Loss}\\ 
    \hline
    1 & 0.795 & 0.72933 & 1 & 0.9656 & 0.90671\\
    \cline{4-6}
     &  &  & 2 & 0.8993 & 0.841933\\
    \hline
    2 & 0.6802 & 0.626306 & 3 & 0.8356 & 0.796878\\
    
    \cline{4-6}
     &  &  & 4 & 0.7891 & 0.766159\\
    \hline
    3 & 0.627 & 0.577756 & 5 & 0.7757 & 0.749898\\
    \hline
  \end{tabular}
  
  \label{table_label9}
\end{table*}

\begin{figure*}[htbp]
    \centering
    \includegraphics[width=\textwidth]{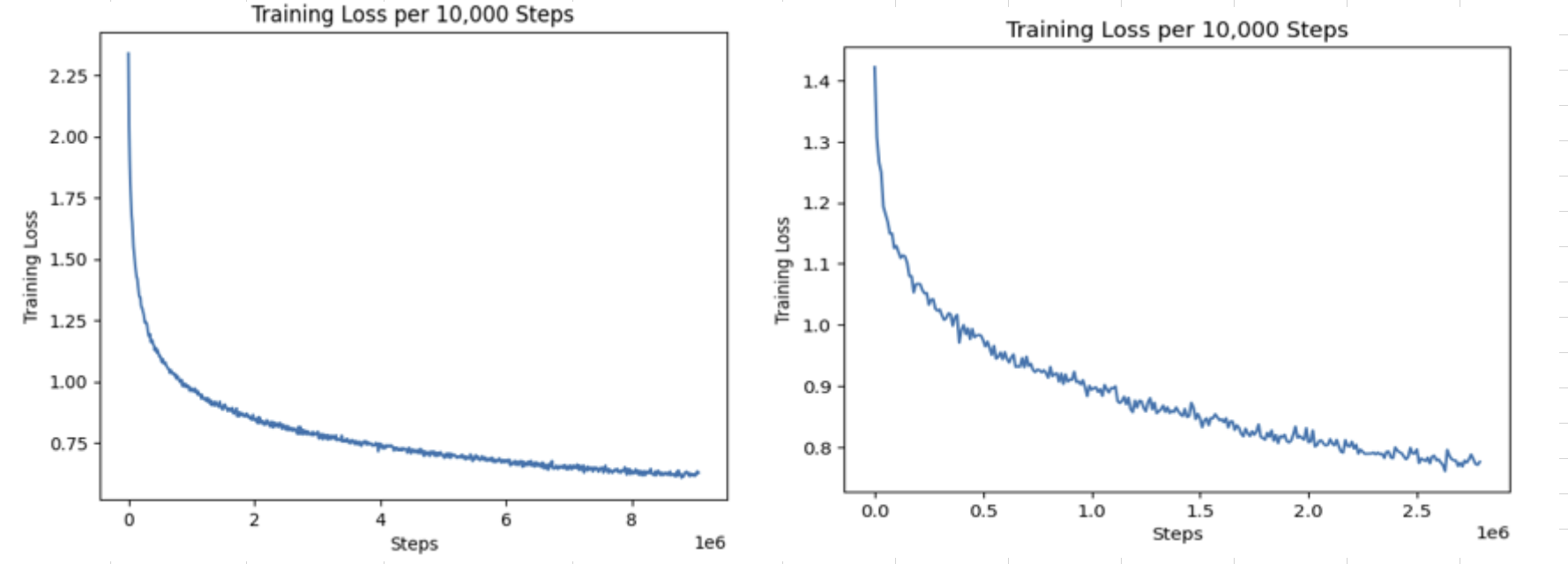}
    \caption{Training Loss: Tri-Distil-BERT (left), Mixed-Distil-BERT (right)}
    \label{fig:your-figure3}
\end{figure*}

For the pretraining, the Environment we have used is Google Colab, GPU Type Nvidia A100, GPU Memory 40 GB, System Memory 83.5 GB, Disk 166.8 GB, Number of Processor is 4 and 16 for Tri-Distil-BERT and Mixed-Distil-BERT respectively. We have used OSCAR Hindi and OSCAR Bangla datasets for pretraining Tri-Distil-BERT and custom 560k code mixed data generated from Yelp polarity by random code-mixing algorithm \cite{krishnan2021multilingual} for Mixed-Distil-BERT.\\
The parameters used as follows: train-test 80-20, mlm\_probability 0.15, per\_device\_train\_batch\_size 16, save\_total\_limit 	3, evaluation\_strategy epoch, learning\_rate 5.00E-05, weight\_decay 0.01 and num\_train\_epochs 3 and 5 for Tri-Distil-BERT and Mixed-Distil-BERT respectively.\\
The total pretraining time for Tri-Distil-BERT is 18 Hours 35 Minutes and for Mixed-Distil-BERT 5 Hours 49 Minutes.\\
The training statistics of the pre-trained Tri-Distil-BERT and Mixed-Distil-BERT are available in table \ref{table_label9}, training loss graphs of the pre-trained models are available in figure \ref{fig:your-figure3}, and perplexity score for our pre-trained models is available in table \ref{table_label10}.

\begin{table*}
  \centering
  \footnotesize
  \setlength{\tabcolsep}{3.75pt}
  \renewcommand{\arraystretch}{1.5}
  \caption{Perplexity Score of our Pre-trained Models}
  \begin{tabular}{ | c | c | c | c | }
    \hline
    \multicolumn{2}{|c|}{\textbf{Tri-Distil-BERT}} & \multicolumn{2}{|c|}{\textbf{Mixed-Distil-BERT}} \\
    \cline{1-4}
    \textbf{Epoch} & \textbf{Perplexity Score} & \textbf{Epoch} & \textbf{Perplexity Score}\\ 
    \hline
    1 & 2.07 & 1 & 2.48\\
    \cline{3-4}
     &  & 2 & 2.32\\
    \hline
    2 & 1.87 & 3 & 2.22\\
    
    \cline{3-4}
     &  & 4 & 2.15\\
    \hline
    3 & 1.78 & 5 & 2.12\\
    \hline
  \end{tabular}
  
  \label{table_label10}
\end{table*}

\begin{table*}
  \centering
  \footnotesize
  \setlength{\tabcolsep}{3.75pt}
  \renewcommand{\arraystretch}{1.5}
  \caption{Emotion detection of 3 languages code-mixed dataset}
  \begin{tabular}{ | c | c | c | c | c | c |}
    \hline
    & \multicolumn{5}{c|}{\textbf{Synthetic Dataset}} \\
    \cline{2-6}
    & \textbf{Accuracy} & \textbf{Weighted F1-Score} & \textbf{Precision} & \textbf{Recall} & \textbf{Time (Min)} \\ 
    \hline
    \textbf{DistilBERT} & 0.43 & 0.40 & 0.42 & 0.43 & 32 \\
    \hline
    \textbf{BERT} & 0.46 & 0.44 & 0.45 & 0.46 & 39 \\
    \hline
    \textbf{roBERTa} & 0.43 & 0.41 & 0.42 & 0.43 & 62 \\
    \hline
    \textbf{BanglaBERT} & 0.41 & 0.39 & 0.40 & 0.41 & 42 \\
    \hline
    \textbf{HindiBERT} & 0.39 & 0.32 & 0.29 & 0.39 & 47 \\
    \hline
    \textbf{mBERT} & 0.50 & 0.49 & 0.49 & 0.50 & 94 \\
    \hline
    \textbf{XLM-R} & \textbf{0.52} & \textbf{0.51} & \textbf{0.51} & \textbf{0.52} & \textbf{116} \\
    \hline
    \textbf{BanglishBERT} & 0.48 & 0.47 & 0.47 & 0.48 & 83 \\
    \hline
    \textbf{IndicBERT} & 0.43 & 0.38 & 0.42 & 0.43 & 64 \\
    \hline
    \textbf{HingBERT} & 0.47 & 0.45 & 0.46 & 0.47 & 68 \\
    \hline
    \textbf{MuRIL} & 0.46 & 0.45 & 0.48 & 0.47 & 36\\
    \hline
    \textbf{EmoBERTa} & 0.43 & 0.41 & 0.42 & 0.43 & 73 \\
    \hline
    \textbf{Tri-Distil-BERT} & \textbf{0.49} & \textbf{0.48} & \textbf{0.48} & \textbf{0.49} & \textbf{33} \\
    \hline
    \textbf{Mixed-Distil-BERT} & \textbf{0.51} & \textbf{0.50} & \textbf{0.50} & \textbf{0.51} & \textbf{36} \\
    \hline
  \end{tabular}
  
  \label{table_label2}
\end{table*}

\begin{table*}
  \centering
  \footnotesize
  \setlength{\tabcolsep}{3.75pt}
  \renewcommand{\arraystretch}{1.5}
  \caption{Sentiment Analysis of 3 languages code-mixed dataset}
  \begin{tabular}{ | c | c | c | c | c | c |}
    \hline
    & \multicolumn{5}{c|}{\textbf{Synthetic Dataset}} \\
    \cline{2-6}
    & \textbf{Accuracy} & \textbf{Weighted F1-Score} & \textbf{Precision} & \textbf{Recall} & \textbf{Time (Min)} \\
    \hline
    \textbf{DistilBERT} & 0.67 & 0.66 & 0.66 & 0.67 & 72 \\
    \hline
    \textbf{BERT} & 0.68 & 0.68 & 0.68 & 0.68 & 96 \\
    \hline
    \textbf{roBERTa} & 0.66 & 0.66 & 0.67 & 0.66 & 118 \\
    \hline
    \textbf{BanglaBERT} & 0.60 & 0.60 & 0.60 & 0.60 & 94 \\
    \hline
    \textbf{HindiBERT} & 0.59 & 0.59 & 0.59 & 0.59 & 102 \\
    \hline
    \textbf{mBERT} & 0.74 & 0.74 & 0.74 & 0.74 & 122 \\
    \hline
    \textbf{XLM-R} & \textbf{0.77} & \textbf{0.77} & \textbf{0.77} & \textbf{0.77} & \textbf{165} \\
    \hline
    \textbf{BanglishBERT} & 0.72 & 0.72 & 0.72 & 0.72 & 132 \\
    \hline
    \textbf{IndicBERT} & 0.65 & 0.66 & 0.67 & 0.65 & 114 \\
    \hline
    \textbf{HingBERT} & 0.67 & 0.67 & 0.67 & 0.67 & 105 \\
    \hline
    \textbf{MuRIL} & 0.77 & 0.77 & 0.77 & 0.78 & 62\\
    \hline
    \textbf{Tri-Distil-BERT} & \textbf{0.69} & \textbf{0.69} & \textbf{0.70} & \textbf{0.69} & \textbf{76} \\
    \hline
    \textbf{Mixed-Distil-BERT} & \textbf{0.70} & \textbf{0.70} & \textbf{0.71} & \textbf{0.70} & \textbf{79} \\
    \hline
  \end{tabular}
  \label{table_label3}
\end{table*}

\begin{table*}
  \centering
  \footnotesize
  \setlength{\tabcolsep}{3.75pt}
  \renewcommand{\arraystretch}{1.5}
  \caption{Offensive language detection of 3 languages code-mixed dataset}
  \begin{tabular}{ | c | c | c | c | c | c |}
    \hline
    & \multicolumn{5}{c|}{\textbf{Synthetic Dataset}} \\
    \cline{2-6}
    & \textbf{Accuracy} & \textbf{Weighted F1-Score} & \textbf{Precision} & \textbf{Recall} & \textbf{Time (Min)} \\
    \hline
    \textbf{DistilBERT} & 0.80 & 0.80 & 0.80 & 0.80 & 32 \\
    \hline
    \textbf{BERT} & 0.83 & 0.83 & 0.83 & 0.83 & 41 \\
    \hline
    \textbf{roBERTa} & 0.80 & 0.79 & 0.79 & 0.80 & 48 \\
    \hline
    \textbf{BanglaBERT} & 0.77 & 0.76 & 0.76 & 0.77 & 40 \\
    \hline
    \textbf{HindiBERT} & 0.75 & 0.73 & 0.74 & 0.75 & 39 \\
    \hline
    \textbf{mBERT} & 0.88 & 0.88 & 0.88 & 0.88 & 62 \\
    \hline
    \textbf{XLM-R} & \textbf{0.88} & \textbf{0.88} & \textbf{0.88} & \textbf{0.88} & \textbf{74} \\
    \hline
    \textbf{BanglishBERT} & 0.86 & 0.86 & 0.86 & 0.86 & 64 \\
    \hline
    \textbf{IndicBERT} & 0.82 & 0.82 & 0.82 & 0.82 & 59 \\
    \hline
    \textbf{HingBERT} & 0.83 & 0.82 & 0.82 & 0.83 & 61 \\
    \hline
    \textbf{MuRIL} & 0.82 & 0.81 & 0.81 & 0.82 & 64 \\
    \hline
    \textbf{fBERT} & 0.81 & 0.81 & 0.81 & 0.82 & 63 \\
    \hline
    
    \textbf{HateBERT} & 0.81 & 0.81 & 0.81 & 0.82 & 67 \\
    \hline
    \textbf{Tri-Distil-BERT} & \textbf{0.86} & \textbf{0.86} & \textbf{0.86} & \textbf{0.86} & \textbf{39} \\
    \hline
    \textbf{Mixed-Distil-BERT} & \textbf{0.87} & \textbf{0.87} & \textbf{0.87} & \textbf{0.87} & \textbf{43} \\
    \hline
  \end{tabular}
  
  \label{table_label4}
\end{table*}

\section{Results and Analysis}

As we mentioned earlier, multi-lingual code-mixing is a very challenging domain and only research on bi-lingual code-mixing has taken place upto now, so we wanted to extend this domain by implementing Tri-lingual code-mixed pre-trained model. We have chosen DistilBERT, which is smaller, faster and similar in performance in comparison to BERT model as our base model to work on. On this model, we have performed pre-training by incorporating Bengali and Hindi corpus of OSCAR dataset and created Tri-Distil-BERT. Further, we pretrained this model with 560k data yelp polarity dataset after code-mixing it by random code-mixing algorithm \cite{krishnan2021multilingual} and created Mixed-Distil-BERT.\\
To justify a pre-trained models performance, it is necessary to compare the model's task-specific fine-tuned results with some existing models fine-tuned on same tasks. We have calculated accuracy, Weighted F1-Score, Precision, Recall but we have comapared performances on the basis of Weighted F1-Score.\\

As we have pre-trained our model on DistilBERT-uncased so we have also fine-tuned DistilBERT for comparison. Moreover, LLMs like BERT \cite{devlin2018bert}, roBERTa \cite{liu2019roberta}, mBERT \cite{devlin2018bert}, XLM-R \cite{conneau2019cross} and MuRIL \cite{khanuja2021muril} are also fine-tuned for the 3 tasks. As we are working on Bengali, Hindi and English code-mixing so comparison with BanglaBERT  \cite{bhattacharjee2021banglabert}, HindiBERT \cite{joshi2022l3cube} and IndicBERT \cite{kakwani2020indicnlpsuite} was also performed. Though we don't have any existing tri-lingual pre-trained model, we use bi-lingual BanglishBERT \cite{bhattacharjee2021banglabert} and HingBERT \cite{nayak2022l3cube} for comparison. Moreover, We use task fine-tuned models EmoBERTa \cite{DBLP:journals/corr/abs-2108-12009} for emotion detection and HateBERT \cite{caselli-etal-2021-hatebert}, fBERT \cite{sarkar-etal-2021-fbert-neural} for offensive language language identification.\\

Among the existing models, for the synthetic data, XLM-R performs the best. Mixed-Distil-BERT has been outperformed by 1\%, 7\%, 1\% by XLM-R on emotion detection, sentiment analysis and offensive language identification for synthetic data. \\




\begin{figure}[htbp]
    \centering
    \begin{tikzpicture}
    \begin{axis}[
        ybar,
        symbolic x coords={HindiBERT, IndicBERT, BanglaBERT, DistilBERT, EmoBERTa, roBERTa, BERT, HingBERT,  MuRIL, BanglishBERT, mBERT, XLM-R, Tri-Distil-BERT, Mixed-Distil-BERT},
        xtick=data,
        x tick label style={rotate=45, anchor=east},
        ylabel=Weighted Weighted F1 Score,
        ymin=0,
        ymax=1,
        width=\linewidth,
        height=6cm,
        bar width=10pt,
        nodes near coords,
        nodes near coords align={vertical},
        ]
        \addplot[fill=blue!50] coordinates {
            (HindiBERT,0.32)
            (IndicBERT,0.38)
            (BanglaBERT,0.39)           
            (DistilBERT,0.40)
            (EmoBERTa,0.41)
            (roBERTa,0.41)
            (BERT,0.44)
            (HingBERT,0.45)
            (MuRIL,0.45)
            (BanglishBERT,0.47)
            (mBERT,0.49)            
            (XLM-R,0.51) 
            (Tri-Distil-BERT,0.48)
            (Mixed-Distil-BERT,0.50)
            
        };
    \end{axis}
    \end{tikzpicture}
    \caption{3 Language Code-Mixed Emotion Detection: Weighted F1-Score Comparison}
    \label{fig:your-figure4}
\end{figure}
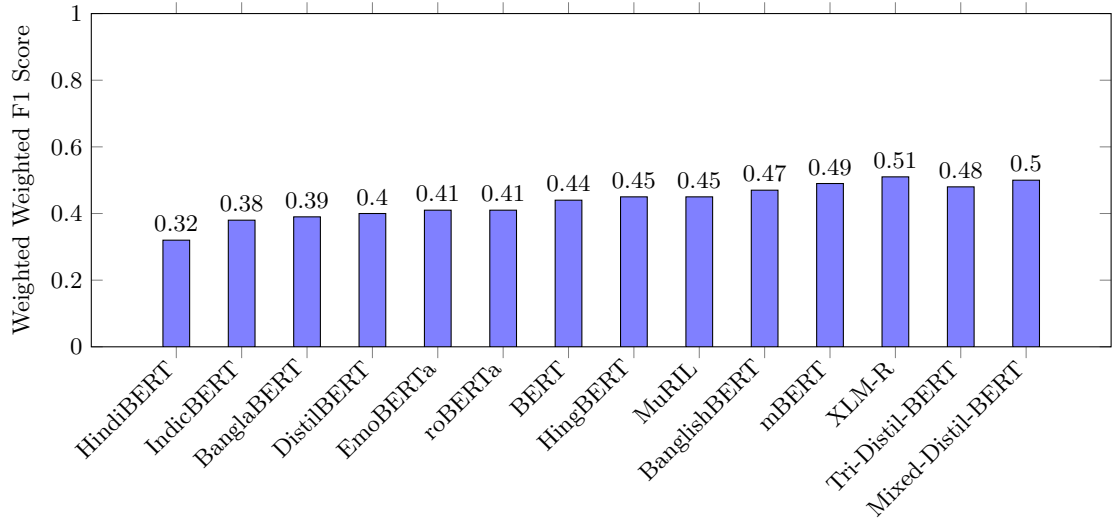

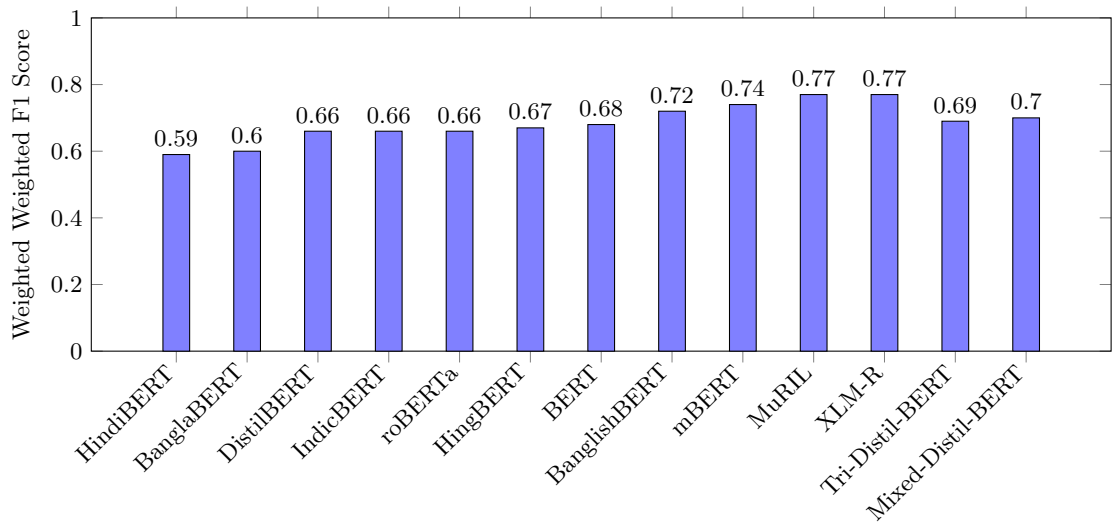
\begin{figure}[htbp]
    \centering
    \begin{tikzpicture}
    \begin{axis}[
        ybar,
        symbolic x coords={HindiBERT, BanglaBERT,  DistilBERT, IndicBERT, roBERTa, HingBERT, BERT, BanglishBERT, mBERT, MuRIL, XLM-R, Tri-Distil-BERT, Mixed-Distil-BERT},
        xtick=data,
        x tick label style={rotate=45, anchor=east},
        ylabel=Weighted Weighted F1 Score,
        ymin=0,
        ymax=1,
        width=\linewidth,
        height=6cm,
        bar width=10pt,
        nodes near coords,
        nodes near coords align={vertical},
        ]
        \addplot[fill=blue!50] coordinates {
            (HindiBERT,0.59)
            (BanglaBERT,0.60)           
            (DistilBERT,0.66)
            (IndicBERT,0.66)
            (roBERTa,0.66)
            (HingBERT,0.67)
            (BERT,0.68)
            (BanglishBERT,0.72)
            (mBERT,0.74)
            (MuRIL,0.77)
            (XLM-R,0.77)
            (Tri-Distil-BERT,0.69)
            (Mixed-Distil-BERT,0.70)
            
        };
    \end{axis}
    \end{tikzpicture}
    \caption{3 Language Code-Mixed Sentiment Analysis: Weighted F1-Score Comparison}
    \label{fig:your-figure5}
\end{figure}

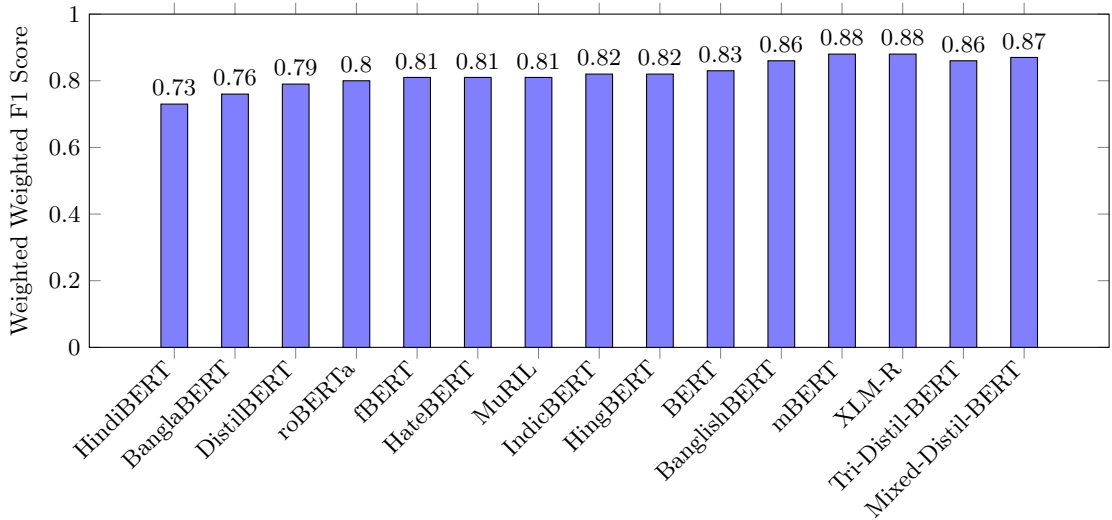
\begin{figure}[htbp]
    \centering
    \begin{tikzpicture}
    \begin{axis}[
        ybar,
        symbolic x coords={HindiBERT, BanglaBERT, DistilBERT, roBERTa, fBERT, HateBERT, MuRIL, IndicBERT, HingBERT, BERT, BanglishBERT, mBERT, XLM-R, Tri-Distil-BERT, Mixed-Distil-BERT},
        xtick=data,
        x tick label style={rotate=45, anchor=east},
        ylabel=Weighted Weighted F1 Score,
        ymin=0,
        ymax=1,
        width=\linewidth,
        height=6cm,
        bar width=10pt,
        nodes near coords,
        nodes near coords align={vertical},
        ]
        \addplot[fill=blue!50] coordinates {
            (HindiBERT,0.73)
            (BanglaBERT,0.76)
            (DistilBERT,0.79)
            (roBERTa,0.80)
            (fBERT,0.81)
            (HateBERT,0.81)
            (MuRIL,0.81)
            (IndicBERT,0.82)
            (HingBERT,0.82)
            (BERT,0.83)           
            (BanglishBERT,0.86)            
            (mBERT,0.88)
            (XLM-R,0.88)
            (Tri-Distil-BERT,0.86)
            (Mixed-Distil-BERT,0.87)
            
        };
    \end{axis}
    \end{tikzpicture}
    \caption{3 Language Code-Mixed Offensive Language Identification: Weighted F1-Score Comparison}
    \label{fig:your-figure6}
\end{figure}

\section{Conclusion}

In summary, we have pre-trained two models Tri-Distil-BERT and Mixed-Distil-BERT for Bengali, Hindi, and English tri-lingual code-mixing for multi-label emotion detection, sentiment analysis, and offensive language identification tasks. We have created 100k synthetic code-mixed data for each of the tasks. \\

Due to computing resources and time constraints, we couldn't use roBERTa model as the base of our pre-training for these tasks. Moreover, the amount of data can be increased for the pretraining of Mixed-Distil-BERT. Last of all, we also have a vision to include more downstream tasks under this research direction.

\bibliography{natl}
\bibliographystyle{acm}

\vspace{2cm}

\section*{Authors}
\noindent {\bf Md Nishat Raihan} Pursuing his Ph.D. in Computer Science at George Mason University, Fairfax, VA, USA. Worked as a Software Engineer at Samsung R\&D for 2 years. Research interests primarily include Code-Mixing, Transliteration, and Large Language Models. Also interested in Time Series Data Mining and Computer Vision Algorithms.\\

\noindent {\bf Dhiman Goswami} Pursuing his Ph.D. in Computer Science at George Mason University, Fairfax, VA, USA. Worked as a lecturer in the Department of Computer Science and Engineering at Daffodil International University, Dhaka, Bangladesh for 3 years. Research interests primarily include Code-Mixing, Native Language Identification, and Large Language Models.\\

\noindent {\bf Antara Mahmud} Completed her M.Sc. in Computer Science at George Mason University, Fairfax, VA, USA. Worked as a lecturer in the Department of Computer Science and Engineering at Daffodil International University, Dhaka, Bangladesh for 3 years. \\

\end{document}